\documentclass[10pt,journal,compsoc]{IEEEtran}
\usepackage{amsmath}
\usepackage{amsfonts, amssymb}
\usepackage{stmaryrd} 
\usepackage{bm}
\usepackage{mathtools}
\usepackage{booktabs}
\usepackage{tabularx}
\usepackage{tabularray}
\usepackage{multirow}
\usepackage{algorithm}
\usepackage[noend]{algpseudocode}
\usepackage{bbm}
\usepackage[backend=bibtex, style=ieee, url=false]{biblatex}
\usepackage{subcaption}
\usepackage{hyperref}
\usepackage{cleveref}
\crefformat{footnote}{#2\footnotemark[#1]#3}
\usepackage{pifont}
\hypersetup{
    colorlinks=true,
    linkcolor=red,
    filecolor=magenta,      
    urlcolor=cyan,
    pdftitle={APN},
    pdfpagemode=FullScreen,
    }
\usepackage{array}
\begin{document}
\title{Progression-Guided Temporal Action Detection in Videos}

\author{Chongkai~Lu,
        Man-Wai Mak, {\it Senior Member, IEEE},
        Ruimin~Li, {\it Member, IEEE},
        Zheru~Chi,  {\it Member, IEEE}, and
        Hong~Fu%
\IEEEcompsocitemizethanks{
\IEEEcompsocthanksitem Chongkai Lu, Man-wai Mak, and Zheru Chi are with the Department of Electrical and Information Engineering, The Hong Kong Polytechnic University, Hong Kong, China.\protect\\
E-mail: \{chong-kai.lu, man.wai.mak, enzheru\}@polyu.edu.hk

\IEEEcompsocthanksitem Ruimin Li is with the Academy of Advanced Interdisciplinary Research, Xidian University, Xi'an Shaanxi Province, China. E-mail: rmli@xidian.edu.cn

\IEEEcompsocthanksitem Hong Fu is with The Education University of Hong Kong, Hong Kong, China. E-mail: hfu@eduhk.hk
}}%


\IEEEtitleabstractindextext{%
\begin{abstract}

We present a novel framework, Action Progression Network (APN), for temporal action detection (TAD) in videos. The framework locates actions in videos by detecting the action evolution process. To encode the action evolution, we quantify a complete action process into 101 ordered stages (0\%, 1\%, ..., 100\%), referred to as action progressions. We then train a neural network to recognize the action progressions. The framework detects action boundaries by detecting complete action processes in the videos, e.g., a video segment with detected action progressions closely follow the sequence 0\%, 1\%, ..., 100\%. The framework offers three major advantages: (1) Our neural networks are trained end-to-end, contrasting conventional methods that optimize modules separately; (2) The APN is trained using action frames exclusively, enabling models to be trained on action classification datasets and robust to videos with temporal background styles differing from those in training; (3) Our framework effectively avoids detecting incomplete actions and excels in detecting long-lasting actions due to the fine-grained and explicit encoding of the temporal structure of actions. Leveraging these advantages, the APN achieves competitive performance and significantly surpasses its counterparts in detecting long-lasting actions. With an IoU threshold of 0.5, the APN achieves a mean Average Precision (mAP) of 58.3\% on the THUMOS14 dataset and 98.9\% mAP on the DFMAD70 dataset. 


\end{abstract}

}

\maketitle

\IEEEraisesectionheading{\section{Introduction}\label{sec:introduction}}


\IEEEPARstart{W}ith the advancement of communication technology, video has become the primary medium for consumer Internet traffic and its proportion is continually increasing \cite{forecast2019cisco}. The rapid growth of video content has led to a growing demand for powerful AI techniques for automatic video understanding, particularly human action recognition. The action recognition community has primarily focused on the task of action classification, which aims to recognize actions in videos that have been trimmed to contain only action content. However, videos are usually unconstrained in practice, containing a significant amount of temporal background content. Consequently, a new task known as temporal action detection (TAD) has drawn increasing attention. TAD requires detecting both the categories (classification) and the temporal boundaries (localization) of actions in untrimmed videos, which may contain multiple action instances from different classes and substantial temporal background.

Nevertheless, an untrimmed video is too large to be directly input into an artificial neural network. To overcome this problem, some studies \cite{oneata2013action, oneata2014lear, wang2014action, shou2016temporal} employ the intuitive "sliding window" paradigm to crop short videos from the untrimmed video and assess the action presence in each short video.  However, sliding window based methods are known to has a tension between computational tractability and high detection quality \cite{hosang2015makes, gao2018ctap}. As a result, \textit{bottom-up} style TAD methods have become prevalent in recent years \cite{singh2016untrimmed, yuan2017temporal, zhao2017temporal, Lin_2018_ECCV, lin2019bmn, liu2019multi, zeng2019graph, zhao2020bottom, gong2020scale, lin2020fast, xu2020g}. For instance, \cite{singh2016untrimmed} evaluates action presence for individual video frames and then groups continuous snippets with high confidence scores as action proposals. In a subsequent study, \cite{yuan2017temporal} detects action presence and encodes localization information by classifying video snippets into three action evolution stages: {\it starting}, {\it course}, and {\it ending}.

\begin{figure*}[ht!]
	\centering
	\includegraphics[keepaspectratio, width=1\textwidth]{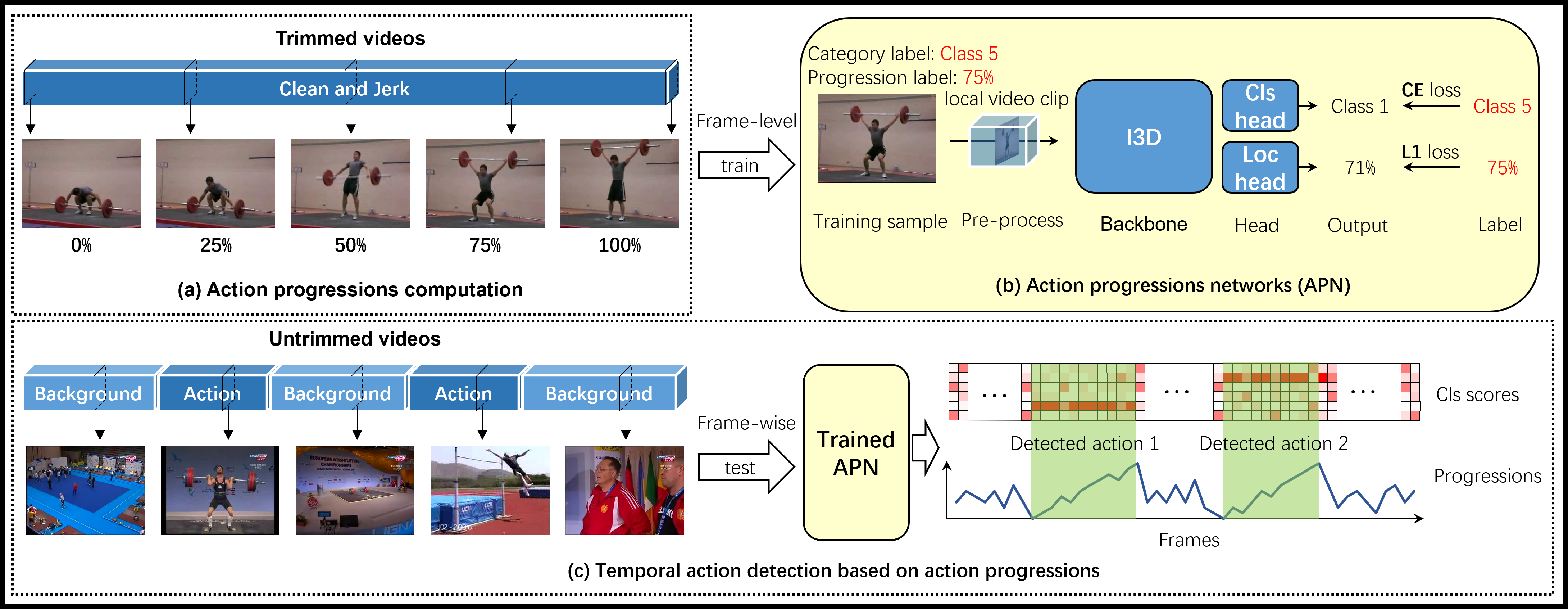}
	\caption{Overview of our framework. (a) Action progression labels (in \%) are first generated for each action frame, based on the relative temporal location of the frames within the action. (b) An artificial neural network, dubbed action progression network, is then trained to predict the action progressions and categories from action frames. Each action frame is extended to include its neighboring frames before being fed to the model. (c) Using the trained APN, we predict the action progression sequence of a test video. A profile-matching algorithm is then applied to the sequence to detect the temporal action boundaries and classification scores.}
	\label{fig: frame_work}
\end{figure*}

Current bottom-up TAD methods classify the three action evolution stages without considering their order. We argue that classifying the entire action into a single \textit{course} stage is imprecise and does not fully utilize the distinct evolution pattern inside the actions. Therefore, we propose a new approach to quantitatively encode action evolution by dividing a complete action into many ordered ranks, termed \textit{action progressions}. As an illustrative example in Fig.~\ref{fig: frame_work}(a), we divide the action evolution into 101 contiguous stages (0\%, 1\%, ..., 100\%) . In contrast to other bottom-up methods with three discrete stages, action progressions encode action evolution with fine granularity to support precise action detection. More importantly, action progressions are represented by numerical values rather than categorical classes; thus, encapsulating the order among evolution stages leads to a better representation of the action evolution.

In addition to the precise encoding of the action evolution, action progressions enable a streamlined TAD framework. Most existing TAD models \cite{gao2017turn, gao2018ctap, Lin_2018_ECCV, lin2019bmn,liu2019multi,lin2020fast, xu2020g, zhao2020bottom, zeng2019graph, jain2020actionbytes, sstad_buch_bmvc17, liu2022end} detect actions by taking as input the entire untrimmed video; thus, they must first perform offline frame/clip-level feature extraction for the untrimmed video. We argue that the offline feature extraction not only complicates the workflow but also limits the detection performance, as the offline features cannot be further fine-tuned for temporal action detection. Thanks to the proposed action progressions, our models only take as input a short video segment, e.g., 4 seconds, from the untrimmed video, as shown in Fig.~\ref{fig: frame_work}(b). This allows our model to accept raw videos as input. Furthermore, in our models, action classification and localization are achieved using two sibling heads sharing a common backbone. The integration of feature extraction, action localization, and action classification enables our networks to be trained in an end-to-end manner. Moreover, our models are trained exclusively on action frames, making them more flexible in terms of training data requirements, e.g., traning on action classification datasets \cite{jain2020actionbytes}.

We refer to the models that predict action progressions as Action Progression Networks (APNs). Fig.~\ref{fig: frame_work} illustrates the processing pipeline of the framework and demonstrates how an APN addresses the TAD problem. The proposed framework first determines an action progression for each action frame based on its temporal location within the action (Fig.~\ref{fig: frame_work}(a)). Then, it trains a neural network to predict the action progressions and categories from the action frames (Fig.~\ref{fig: frame_work}(b)). We also utilize surrounding frames to provide additional temporal information to the network (local video clip in Fig.~\ref{fig: frame_work}(b)). A critical aspect of the APN is that its localization head predicts action progressions by solving an ordinal regression problem rather than a classification problem. Framing the identification of the progression pattern as a regression problem improves the precision of predicting
action progressions (see Table~\ref{tab: heads_compare}).

During inferencing, the APN locates the temporal action boundaries in an untrimmed video by analyzing the pattern of action progressions (Fig.~\ref{fig: frame_work}(c)). Because APNs are trained exclusively on the action frames, the predicted progressions of the background activities are noisy, while the progressions of foreground actions exhibit a distinct pattern. More precisely, the values of the predicted progressions of a complete action increase almost linearly from 0\% to 100\%. With such distinct difference between the patterns of actions and background, the framework can locate the actions in videos using a simple profile-matching algorithm. More importantly, since action progressions can be predicted for every frame in a test video, the framework can locate the actions at a high temporal resolution, i.e., at the frame level. As for the classification, we simply average the class scores across the frames within an action, which are predicted by the classification head in the APN.



For the APN's localization head, we studied five representative ordinal regression methods for predicting the action progressions, including two primary forms -- nominal classification and regression -- and three advanced schemes -- cost-sensitive classification, binary decomposition, and the threshold model. We discovered that the performance of APNs is sensitive to the choice of the ordinal regression methods, and all methods outperform the \textit{nominal classification} baseline. Our results validate the importance of the order of action evolution stages. We also investigated the impact of varying the dissection resolution (the number of ranks in a progression pattern) and the number of sampled frames in test videos on detection performance. It shows that increasing the dissection resolution and the number of frames sampled improve detection performance. These result validate the effectiveness of applying fine-grained dissection to actions and videos. Additionally, we examined the impact of APN's input duration and found that increasing the input's time span results in higher detection precision, which is expected, as a longer temporal input contains richer dynamic information.




We conducted experiments on two challenging temporal action detection datasets, including the THUMOS14 \cite{jiang2014thumos} and a new dataset, DFMAD70, created by ourselves. The APN achieves competitive performance on THUMOS14 (mAP@0.5=58.3\%) under the condition that our models are end-to-end and trained on action frames only. Note that another TAD method \cite{jain2020actionbytes} that also uses trimmed videos during training can only achieve mAP@0.5=15.5\%. In DFMAD70, whose average duration of actions is much longer than THUMOS14, our APN significantly outperforms other detectors. In particular, compared to the PGCN \cite{zeng2019graph} that uses proposals from G-TAD \cite{xu2020g}, the APN achieves a superior detection precision (96.0\% vs. 77.9\%) with a 30 times faster inference speed. This outstanding performance is attributed to our framework's ability to leverage long-duration actions through fine-grained dissection of actions. The experimental results on DFMAD70 also demonstrate the APN's ability to avoid detecting incomplete actions.


Overall, we propose quantifying action evolution for temporal action detection and make \textit{three contributions}: 1) we introduce a new characteristic of action frames, dubbed action progressions, to quantitatively encode the temporal evolution of actions; 2) Based on the proposed action progressions, we design an effective yet simple temporal action detection framework that is trained end-to-end on action frames only; and 3) The proposed framework demonstrates outstanding performance in detecting long-duration actions, and to the best of our knowledge, it is the first TAD method that explicitly avoids detecting incomplete actions.

This paper extends our previous work \cite{lu2021precise} in several aspects. \textit{First,} we use a stack of adjacent frames instead of a single frame as input to the APN, which improves the accuracy of action progression prediction. \textit{Second,} we employ a single model for all action classes rather than training one model for each class. \textit{Third,} We utilize more advanced loss functions and neural network designs. \textit{Fourth,} we conduct a more extensive range of experiments and analyses, such as evaluating the APN on the THUMOS14 dataset to demonstrate its superior performance.

\section{Related Work}
\subsection{Temporal Action Detection}
Temporal action detection (TAD) is a meaningful yet challenging task. To avoid the extensive cost of annotations, some approaches \cite{sun2015temporal, singh2017hide, Wang_2017_CVPR, shou2018autoloc, nguyen2018weakly, liu2019weakly, narayan20193c, min2020adversarial, shi2020weakly, luo2020weakly, zhai2020two, qu2021acm, lee2021weakly} tackle this task in a weakly supervised manner by leveraging easily available labels (such as video-level categorical labels). Although weakly supervised methods reduce the burden of labeling temporal annotations, their detection performance is not satisfactory. On the other hand, fully supervised TAD has higher interpretability and performance but requires annotations of both the temporal boundaries and categories of actions in untrimmed videos. Our work is of this type.





Inspired by the success of two-stage object detections \cite{girshick2014rich, girshick2015fast, ren2015faster, he2017mask, Lin_2017_CVPR}, the two-stage architecture has become predominant in the TAD community. The architecture comprises a proposal generator and a proposal recognizer. The former detects temporal boundaries of actions and the latter classifies the detected proposals. Many two-stage TAD methods focus on proposal generation and adopt off-the-shelf action classification models as their proposal recognizers. The commonly used proposal recognizers include: (a) UntrimmedNet \cite{Wang_2017_CVPR} which provides video-level classification scores, (b) SCNN \cite{shou2016temporal} which predicts proposal-level classification scores, and (c) P-GCN \cite{zeng2019graph} which evaluates proposals by analyzing the relationship among the proposals.

There are two main styles of proposal generation. The first style use sliding windows of various temporal length to crop a set of video segments and then computes some scores indicating the likelihood of the cropped segments being an action. For simplicity, we call this style \textit{slide windows ranking}. The second style typically derives some low-level characteristics from the videos, e.g., frame-level \textit{actionness scores} \cite{singh2016untrimmed, zhao2017temporal}, and then generate proposals by analyzing their patterns, e.g., \textit{temporal actionness grouping} \cite{singh2016untrimmed, zhao2017temporal}. We call this style \textit{bottom-up proposal generation.}



The early methods \cite{oneata2013action, Oneata_2014_CVPR, oneata2014lear, wang2014action} of sliding windows ranking are simple: they take all sliding windows as action candidate (proposals) and rank them by the classification scores from the proposal recognizer. Their detection results are imprecise because the localization scores were not evaluated. The later methods evaluate the scores of sliding windows and take the windows with high scores as proposals. For example, Heilbron \textit{et al.} \cite{Heilbron_2016_CVPR} used Spatio-Temporal Interest Points (STIPs) \cite{laptev2005space} encoded by Histogram of Oriented Gradients (HOG) \cite{dalal2005histograms} and Histogram of Optical Flow (HOF) \cite{laptev2008learning} to learn a dictionary for efficient proposal generation. Shou \textit{et al.} \cite{shou2016temporal} trained a 3D ConvNet in a multi-stage manner to learn the generation, classification, and localization of proposals. Escorcia \textit{et al.} \cite{escorcia2016daps} leveraged the vast capacity of deep networks to hypothesize proposals at different scales in a single model. Similarly, Buch \textit{et al.} \cite{Buch_2017_CVPR} utilized the memory of the Gated Recurrent Unit (GRU) to determine multi-scale proposals in a single pass. Based on this idea, they proposed an end-to-end single-stream temporal action detection model \cite{sstad_buch_bmvc17}. Yueng \textit{et al.} \cite{yeung2016end} used reinforcement learning to refine the action boundaries. Gao \textit{et al.} \cite{gao2017turn} trained a neural network to predict two regression offsets, in addition to the confidence scores, to refine the proposal boundaries. Bai \textit{et al.} \cite{bai2020boundary} employed graph neural networks (GNN) to model the relationship between the boundary and action content to locate actions.

Sliding windows ranking methods require evaluating a large number of video segments, which is computationally expensive. In contrast, the proposal generation in bottom-up style can detect actions in fine granularity at a low computation cost through evaluating low-level video characteristics. For example, Gaidon \textit{et al.} \cite{5995646, 6487513} modeled an action as a sequence of key atomic action units (actoms) and detected actions based on their temporal structure. Ma \textit{et al.} \cite{Ma_2016_CVPR} modeled the cumulative activity progressions in videos and combined them with the frame-level classification scores to detect actions. Singh \cite{singh2016untrimmed} directly trained a ConvNet to predict the frame-level action classification scores and used a delicate post-processing algorithm for detection. Yuan \textit{et al.} \cite{yuan2017temporal} predicted the starting, course, and ending scores of every frame and then searched for the frame sequence that has the highest summation scores. Shou \textit{et al.} \cite{Shou_2017_CVPR} proposed a convolutional-deconvolutional network (CDC) that outputs predictions with the same temporal length as the input videos, thereby enabling the network to learn fine-grained temporal contexts. Zhao \textit{et al.} \cite{zhao2017temporal} devised the temporal actionness grouping (TAG) to generate proposals by clustering frames based on the frame-level classification scores. They used a pyramid model to classify the generated proposals by considering their completeness. 

In 2018, Lin \textit{et al.} \cite{Lin_2018_ECCV} proposed predicting the probabilities of action boundaries in video sequences. Later, they improved their work by presenting a novel boundary-matching confidence map to encode the proposals' locations and durations and used them to evaluate the proposals in an end-to-end pipeline \cite{lin2019bmn}. A dense boundaries generator \cite{lin2020fast} was proposed to improve \cite{Lin_2018_ECCV} by combining the proposal generation and evolution in a single stream using confidence maps as the learning target. The authors also introduced a new action-aware stream. Similarly, Liu \textit{et al.} \cite{liu2019multi} used frame actionness to refine the predicted action boundaries. Zhao \textit{et al.} \cite{zhao2020bottom} modeled the temporal constraints between different action phases with two regularizations: intra-phase consistency and inter-phase consistency. Gong \textit{et al.} \cite{gong2020scale} used multiple streams of stacked temporal convolution blocks with different dilation to deal with variable duration of actions. Xu \textit{et al.} \cite{xu2020g} captured the semantic context in videos with graph convolutional networks and cast proposal generation as a sub-graph localization problem. As a supplement, the method proposed by Gao \textit{et al.} \cite{gao2018ctap} takes the sliding window ranking and actionness scores as complementary cues to generate action proposals. Inspired by the DETR \cite{carion2020end}, Liu \textit{et al.} \cite{liu2022end} applied the transformer architecture to solve the proposal generation and classification in one model.

Bottom-up proposal generation is computationally friendly; therefore, they can produce proposals of more flexible location and length. Depending on the features and patterns, bottom-up methods can have similar and even better performance compared with sliding window-based methods.

It's worth to note that many TAD authors claimed their models as end-to-end; however, many of them \cite{liu2022end, sstad_buch_bmvc17} actually only trained the proposal generator and proposal classifier end-to-end, and the feature extractor is treated offline. We argue that such design is not truly end-to-end and the feature extracted from the classification models may be limited when it comes to localizing actions.



\subsection{Ordinal Regression}
In our framework, predicting action progressions from video frames is framed as an ordinal regression problem. A good survey about ordinal regression methods can be found in \cite{ordinal_survery}. According to the proposed taxonomy in \cite{ordinal_survery}, there are five main ordinal regression methods. Among them, three are naive methods: \textit{regression}, \textit{nominal classification}, and \textit{cost-sensitive classification}, and two are advanced methods -- \textit{ordinal binary decomposition} and \textit{threshold models}. We have investigated all of these methods for the APN. The \textit{regression} and \textit{nominal classification} are implemented by ourselves, whereas the others are implemented with references to the existing work. In particular, we implemented \textit{ordinal binary decomposition} by referring to \cite{Niu_2016_CVPR}, \textit{cost-sensitive classification} by referring to \cite{diaz2019soft}, and \textit{threshold models} by referring to \cite{cao2019rank}.

\section{Action Progression Networks}\label{sec:action progression networks}
In this work, rather than detecting actions in videos directly, we used neural networks to tackle a pretext task -- \textit{predicting the action evolution process in videos}. The action localization is accomplished by analyzing the predicted action evolution with a simple algorithm. Since the algorithm is independent of the neural networks, its description is left to Section~\ref{sec: detection in videos}. This section describes how we train a network to recognize the action evolution in videos. Specifically, we first explain how the action progressions are quantitatively encoded as regression labels of action evolution. After that, we present the architecture of the network (called the action progression network (APN)) that predicts action progressions and categories from action frames. Finally, we give the details of five ordinal regression methods for encoding the action progressions.


\subsection{Action Progressions in Videos}\label{sec: action progression}
The majority of deep-learning-based action recognition methods treat the whole action as a learning unit. However, we observer that most action instances of the same type share a similar evolution process. This suggests that there are explicit temporal structures inside the actions, which can be utilized for recognition. Some previous studies \cite{zhao2017temporal, yuan2017temporal, Lin_2018_ECCV, lin2019bmn} have demonstrated the effectiveness of using action evolution for improving detection performance. However, none of them have built a fine-grained encoding scheme for action evolution. For example, \cite{zhao2017temporal} and \cite{yuan2017temporal} split the temporal process of actions into three phases only and did not consider their order during training.

To encode action evolution in a fine-grained style and to take their order into account, we quantify the action process on a numerical scale. As shown in Fig.~\ref{fig: frame_work}(a), we represent the time points in an action evolution as real numbers in the range 0\% to 100\%. By doing this, we can build fine-grained temporal structures from actions, thereby utilizing the patterns of action evolution exhaustively. We call these regressive characteristics of action evolution as \textit{action progressions}.

It's worth noting that the ground truths of action progressions can be derived without human annotation. Specifically, we derive the progression label for every action frame based on their chronological positions in the action instance. For the TAD dataset with annotations in untrimmed videos, we compute the action progressions based on the annotations of temporal boundaries of actions. As for the action classification datasets, we directly derive the action progressions since the videos are trimmed.



Formally, given an action instance comprising $l$ frames $A= \{f_{1}, f_{2}, ..., f_{l}\}$, we assign every frame $f_{\tau}$ with an action progression label $p_{\tau}$ using the following formula:
\begin{equation}
    p_{\tau} = \left\lfloor K\frac{\tau}{l} \right\rceil,
    \label{eq: label equation}
\end{equation}
where $\lfloor \cdot \rceil$ denotes the rounding-to-the-nearest integer, and $K$ is a constant controlling the number of ranks used to dissect the action evolution. In practice, we found that a large $K$ is good for learning fine-grained action evolution. However, increasing $K$ beyond 100 will lead to diminishing returns and requires excessive computation, see Fig.~\ref{fig: pef vs. ranks}. Therefore in our experiments, we fixed $K$ to 100, leading to 101 action progressions, i.e., 0\%, 1\%, ..., 100\%.

For the action progression derived by Eq.~\ref{eq: label equation} to be reasonable and informative, some conditions must be met. That is, the action should have an exact, rich, and non-repetitive evolution pattern, which we call \textit{progressive actions}. In other words, our method is not suitable for detecting non-progressive actions, e.g., the indeterminate actions like \textit{painting}, the ephemeral actions like \textit{hugging}, and the repetitive actions like \textit{walking}. The reward is that our work can detect progressive actions precisely. Besides, we believe non-progressive actions are naturally hard to detect. We leave the discussion about this to the experiment section, Sec.~\ref{sec: experiments}


\subsection{Neural Network Architecture}\label{subsec: model architecture}
We use a deep-learning model to predict the action progressions and categories from video frames. Notably, the APN is different from conventional action detection models. As shown in Fig.~\ref{fig: frame_work}(b), the APN handles one time-point (video frame) in the video at a time, instead of processing a video segment cropped by a sliding window or the features of the whole video. In particular, the APN predicts the action progression at one time point for each forward pass.

\textbf{Input}: Although the APN aims to predict the action progression of each frame, the prediction will be more accurate if the APN receives contextual frames (instead of a single frame) as input. Formally, the input volume $I_{\tau}$ of a frame $f_{\tau} \in\mathbb{R}^{C\times W\times H}$ at time $\tau$ is constructed as follows:
%
\begin{equation}\label{eq: input}
    I_{\tau} = \{f_{\tau-(l-1)d}, \ldots, f_{\tau-d}, f_{\tau}, f_{\tau+d},\ldots ,f_{\tau+ld}\},
\end{equation}
where $d$ is the temporal stride (in frames) between the adjacent frames of the $L=2l$ frames contained in the volume, and $C$, $W$, and $H$ are the dimensions of the channels, width, and height of the videos, respectively. Specifically, $C$ is equal to 3 and 2 when the frames are RGB images and optical flows, respectively. During each training iteration, the APN was trained to predict the progression label $p_{\tau}$ with the input $I_{\tau}$.

\textbf{Backbone}: We studied the I3D \cite{Carreira_2017_CVPR} and the ResNet-50 \cite{he2016deep} as the APN backbone to produce the feature vectors. The ResNet-50 was used when the input is a single 2-D image, i.e., $I_{\tau} = f_{\tau}$; otherwise the I3D was used. The output features of the I3D and the ResNet-50 backbone after average pooling are of size 1024 and 2048, respectively.

It is worth noting that the choice of backbones for the APN is flexible. Thus, it is possible to incorporate the APN with any other advanced models that are good at capturing video dynamics. Here, we only investigate the commonly used backbones because our innovation lies in the workflow rather than the network architecture.

\textbf{Sibling Heads}: The APN has two heads, the classification head and the localization head, referring to as \textit{Cls} and \textit{Loc} head in Fig.~\ref{fig: frame_work}. The classification head is a fully connected layer outputting the classification scores trained with cross-entropy loss. The localization head is designed to output the action progressions. We investigated five different ordinal regression methods for encoding the action progressions, resulting in five types of localization heads. The details of them are described in Section~\ref{sec: ordinal regression methods}. Formally, the sibling heads output the predicted classification scores, $\hat{\bm{s}} \in \mathbb{R}^C \cap (0, 1)$, and a predicted action progression $\hat{p} \in \mathbb{R} \cap [0, K]$, where $C$ and $K$ are the number and action classes and progression ranks, respectively.


\subsection{Ordinal Regression Heads}\label{sec: ordinal regression methods}

From the perspective of the learning target's format, predicting the action progressions, which are numerical and ordered, is an ordinal regression problem. According to the research in ordinal regression \cite{ordinal_survery, Niu_2016_CVPR, diaz2019soft, cao2019rank}, letting the models learn from well-encoded regression targets can result in considerable performance improvement. The encoding schemes are called ordinal regression and are reflected by different output formats and loss functions.


To find a proper setting, we have investigated five representative ordinal regression methods for building the localization head of APN, including the nominal classification, regression, binary decomposition \cite{Niu_2016_CVPR}, cost-sensitive classification \cite{diaz2019soft}, and threshold model \cite{cao2019rank}.

In the following, we suppose that an input volume (Eq.~\ref{eq: input}) with the ground truth action progression label $p \in \mathbb{Z} \cap [0, K]$ is input to the APN. Then, the APN uses different types of heads to learn the action progressions (according to the regression methods adopted).


\textit{\textbf{Regression}}. The regression head is trained to predict the target action progression without any encoding except for a normalization process. The mean absolute error (L1) is used as the loss function:
\begin{equation}\label{eq: reg_loss}
    \mathcal{L}^{1}(o,p) = \left| \mathcal{G}({o}) - \frac{p}{K} \right|,
\end{equation}
where $o \in \mathbb{R}$ is the head's output and $\mathcal{G}$ is a clamp function that normalizes the predicted rank to the range $[0, 1]$. During inferencing, the predicted rank $\hat{p}$ is decoded from the inference output $\hat{o}$ based on the following formula: 
\begin{equation}\label{eq: reg_inf}
    \hat{p}^{1} = K \mathcal{G}(\hat{o}).
\end{equation}

Humans can easily interpret the regression head's outputs, but it may be difficult for machines. The regression head tends to focus on the magnitude rather than ordering \cite{harrington2003online}. Unfortunately, there is no principled way of deciding the regression values \cite{ordinal_survery}. Even worse, The regression labels do not take into account that an action could progress in a different way during evolution, which is known as the \textit{non-stationary} problem \cite{Niu_2016_CVPR}. For example, the evolution of the action ``high jump'' appears as changes in the running speed during its first half, while it appears as changes in the jumping height during the second half.

\textit{\textbf{Nominal Classification}}. Strictly speaking, nominal classification is not an ordinal regression method because it ignores the continuity between the ranks. We put it here as a baseline for consistency. The head simply predicts the action progression as solving a classification problem. Because there are $K+1$ ranks, the output of this head is a vector, $\bm{o} \in \mathbb{R}^{K+1}$. The widely used classification loss, cross-entropy (CE), is used:
\begin{equation}\label{eq: cls_loss}
    \mathcal{L}^{2}(\bm{o}, p) = -\log(\text{softmax}(\bm{o})_{p}) = -\log{\frac{\exp{(o_{p})}}{\sum_{j=0}^{K}\exp{(o_{j})}}}.
\end{equation}

As for inference, because ranks are ordinal, in addition to the conventional arg-max function:
\begin{equation}\label{eq: cls_inf1}
     \hat{p}^{2} = \operatorname*{arg\,max}_{j=0} ^{K} \hat{o}_{j},
\end{equation}
we also investigate the expectation function to decode the prediction result from the head's outputs:
\begin{equation}\label{eq: cls_inf2}
    \hat{p}^{2} = \sum_{j=0}^{K} j\times \text{softmax}(\hat{\bm{o}})_{j}.
\end{equation}
In our experiments, using the expectation function (Eq.~\ref{eq: cls_inf2}) always leads to lower prediction errors compared to the arg-max function (Eq.~\ref{eq: cls_inf1}); thus, it was used as the default setting.

Although the nominal classification head does not have the \textit{non-stationary} problem, it ignores the rank order. For example, the classification losses of predicting rank-1 as rank-2 is the same as predicting rank-1 as rank-99, which is obviously not desirable.

\textit{\textbf{Cost-sensitive Classification}}. This head shares the same output format with the nominal classification head, i.e., $\bm{o} \in \mathbb{R}^{K+1}$. The difference is that it uses soft labels, instead of the one-hot labels, as the learning targets to encode the ordinal relationship among adjacent classes. In our experiment, we used the following formula to convert the original regression label $p$ to its soft label $\bm{q} \in \mathbb{R}^{K+1}$:
\begin{equation}\label{eq: soft_label}
    \bm{q} = [q_{j}]_{j=0}^{K}, \text{where} \; q_{j} = \frac{\exp(-\sqrt{\left|j - p\right|})} {\sum_{k=0}^{K} \exp(-\sqrt{\left|k - p\right|})}.
\end{equation}
\\
Following \cite{diaz2019soft}, we used the Kullback–Leibler divergence to compute the loss of an output $\bm{o}$ when the soft label is $\bm{q}$:
\begin{equation}\label{eq: soft_loss}
    \mathcal{L}^{3}(\bm{o}, \bm{q}) = \sum_{j=0}^{K} q_{j} \log \frac{q_{j}}{\text{softmax}(\bm{o})_{j}}.
\end{equation}
%


The cost-sensitive classification head shares the same inference formula as the nominal classification head. This ordinal regression method is also known as the soft ordinal vector (SORD) \cite{diaz2019soft}. Soft labels encode the rank order in terms of probabilities. The probability value of each rank, however, is hard to determine without prior knowledge of the problem. In other words, using the simple soft label constructor (Eq.~\ref{eq: soft_label}) and applying it to all ranks of all actions are not that reasonable.

\textit{\textbf{Binary Decomposition}}. This head decomposes the ordinal regression problem into multiple binary classification problems -- ``Is the progression $p$ greater than $j$?'', where $j$ takes value from all possible ranks \cite{Niu_2016_CVPR}. Therefore, the output of the binary decomposition head is $\bm{o} \in \mathbb{R}^{K \times 2}$. The CE loss on each binary classifier is computed, and their mean is taken as the final loss:
\begin{equation}\label{eq: bdc_loss}
    \mathcal{L}^{4}(\bm{o}, p) = -\sum_{j=1}^{K} \llbracket p \geq j\ \rrbracket \mathcal{F}(\bm{o}_{j})_{1} + \llbracket p < j \rrbracket\mathcal{F}(\bm{o}_{j})_{2},
\end{equation}
where the subscripts $1$ and $2$ are the indexes on the nodes of each binary classifier. $\llbracket \cdot \rrbracket$ is a Boolean test, which is equal to 1 if the inner condition is true, and 0 otherwise. $\mathcal{F}$ is the log-softmax activation function. The inference formula is:
\begin{equation}\label{eq: bdc_inf}
     \hat{p}^{4} = \sum_{j=1}^{K} \llbracket \text{softmax}(\hat{\bm{o}}_{j})_{1} \geq 0.5 \rrbracket.
\end{equation}

Binary decomposition converts a regression problem into a series of binary classification sub-problems. This design can help alleviate the \textit{non-stationary} problem. However, the decomposition head does not have any limit on the outputs of different ranks, resulting in the \textit{inconsistency} problem \cite{cao2019rank}. For example, it may answer \textit{Yes} to the sub-problem ``Is $\hat{p} \geq 20$?'', and meanwhile answers \textit{No} to the sub-problem ``Is $ \hat{p} \geq 10$?''. 


\textit{\textbf{Threshold Model}}. This head is created by simply adding $K$ learnable bias terms to the conventional regression head. Each biased result is then transformed, by a sigmoid function, to the probability of one binary classification problem (just like the binary decomposition). Notably, unlike binary decomposition, the threshold model performs binary classification by simply outputting the probability of \textit{Yes} to the question  ``Is $\hat{p} \geq$ rank $j$?''. In this way, the binary cross-entropy (BCE) (rather than CE) is applied to each element of tje output $\bm{o} \in \mathbb{R}^{K}$ to compute the loss:
\begin{equation}\label{eq: thr_loss}
    \mathcal{L}^{5}(\bm{o}, p) = -\sum_{j=1}^{K} \llbracket p \geq j \rrbracket \log \mathcal{A}(o_{j}) + \llbracket p < j \rrbracket \log(1 - \mathcal{A}(o_{j})),
\end{equation}
where $\llbracket \cdot \rrbracket$ is the Boolean test and $\mathcal{A}$ is the sigmoid activation function. The inference formula is:
\begin{equation}\label{eq: thr_inf}
     \hat{p}^{5} = \sum_{j=1}^{K} \llbracket \mathcal{A}(\hat{o}_{j}) \geq 0.5 \rrbracket.
\end{equation}

This method, known as the consistent rank logits (CORAL), was first proposed by Cao \textit{et al.} \cite{cao2019rank} to incorporate the threshold model into deep learning. It uses different thresholds (bias terms) to encode different degrees of dissimilarity between the samples of successive ranks, thus helping to solve the non-stationary problem. Besides, the threshold model can naturally learn bias that decreases monotonically from lower ranks to higher ranks, solving the \textit{inconsistency} problem mentioned in the binary decomposition head. 

\section{Temporal Action Detection with APN}\label{sec: detection in videos}
With the APN framework, the action progressions of video frames can be learned and predicted. However, there remains the question of how to use the predicted action progressions to derive the temporal boundaries of actions in the videos. In this section, we describe how to apply a trained APN for temporal action detection. An overview of the detection workflow is displayed in Fig.~\ref{fig: frame_work}(c).

First, given an untrimmed test video with $T$ frames, we present the frames to the trained APN sequentially to predict a sequence of action progressions $\hat{\bm{p}} \in \mathbb{R}^T$ and a class score matrix $\hat{\bm{S}} \in \mathbb{R}^{T \times C}$: 
\begin{equation}
\begin{split}
    \hat{\bm{p}} &=  \begin{bmatrix}\hat{p}_{1} & \hat{p}_{2} & \dots & \hat{p}_{\tau} & \dots & \hat{p}_T\end{bmatrix} \in \mathbb{R}^T, \\
    \hat{\bm{S}} &=  \begin{bmatrix}\hat{\bm{s}}_{1} & \hat{\bm{s}}_{2} & \dots & \hat{\bm{s}}_{\tau} & \dots & \hat{\bm{s}}_T\end{bmatrix} \in \mathbb{R}^{T \times C},
\end{split}
\end{equation}
where $\hat{p}_{\tau} \in \mathbb{R} \cap [0, K]$ and $\hat{\bm{s}}_{\tau} \in \mathbb{R}^{C} \cap (0, 1)$ are the predicted action progression and class scores of the $\tau$-th frame in the test video. After that, we applied a \textit{profile-matching} algorithm to the progression sequence $\hat{\bm{p}}$ and class score matrix $\hat{\bm{S}}$. According to the definition of action progression, the progression sequence of an ideal complete action is an arithmetic sequence starting with 0 and ending with $K$. Therefore, if any sub-sequences in the progression sequence fit this target pattern, they must be associated with the action instances. Algorithm~\ref{Alg: action search} is designed based on this principle, and it is applied to the progression sequence of the test video $\hat{\bm{p}} \in \mathbb{R}^{T}$ together with the class score matrix $\hat{\bm{S}} \in \mathbb{R}^{T \times C}$ to output the detected actions $\Psi$.
\begin{algorithm}[!h]
	\caption{Action Detection on Progression Sequence}
	\label{Alg: action search}        
	\textbf{Input:} \\
	Progression sequence: $\hat{\bm{p}} \in \mathbb{R}^T$;\\
    Class scores: $\hat{\bm{S}} \in \mathbb{R}^{T \times C}$; \\
	\textbf{Parameters:}\\
	Threshold (Min.) of the action length: $T_{len}^{min}$\\
	Threshold (Max.) of the starting progression:  $T_{start}^{max}$\\
	Threshold (Min.) of the ending progression: $T_{end}^{min}$\\
	Threshold (Min.) of the IoU in NMS: $T_{IoU}$\\
	\textbf{Initialization:} \\
	Detected actions: $\Psi \gets \varnothing$\\    
    $starts = \{\tau \mid \hat{p}_{\tau} < T_{start}^{max}\}$ \\
    $ends = \{\tau \mid \hat{p}_{\tau} > T_{end}^{min}\}$
	\begin{algorithmic}[1]
		\For{$s \; \textbf{in} \; starts$}
		\For{$e \; \textbf{in} \; ends$}
		\State $l \gets e - s + 1$
		\If{$l > T_{len}^{min}$}
		\State $\text{profile} \gets \left[\hat{p}_{\tau}\right]_{\tau=s}^{e}$
		\State $\text{template} \gets \left[K \tau / l\right]_{\tau=0}^{l}$    
		\State $\text{grade} \gets \mathcal{E}(\text{profile}, \text{template})$
		\If{$\text{grade} > 0$}
		\State $\bm{c} = \sum_{i=s}^{e}\hat{\bm{s}}_i/l$
		\State $\Psi \cup \{(s, e, \bm{c}, \text{grade})\} $
		\EndIf
		\EndIf
		\EndFor
		\EndFor
		\State $\Psi = \text{NMS}(\Psi, T_{IoU})$
	\end{algorithmic}
    \textbf{Output:} \\ 
    Detected actions: $\Psi$
\end{algorithm}

In Algorithm~\ref{Alg: action search}, NMS stands for the non-maximum-suppression \cite{rosenfeld1971edge}, which is used to reduce redundant detections; $\mathcal{E}(\cdot)$ reflects the similarity between two profiles of the same length, and it is used for evaluating the detection confidence (grade). In practice, we compute $\mathcal{E}(\cdot)$ based on the Euclidean distance as follow:
\begin{equation} \label{eq: grade}
    \mathcal{E}(\bm{v_1}, \bm{v_2}) = 1 -\frac{\text{MSE}(\bm{v_1}, \bm{v_2})}{1666.66},
\end{equation}
%
where 1666.66 is the mean square error (MSE) of random pairs when $K=100$\footnote{\label{first}The proof can be found in the GitHub page of this paper: \url{https://github.com/makecent/APN}.}. This grading function guarantees that when the two profiles are the same, it will get a full grade of 1, and when the difference between the two profiles is greater than or equal to a random guess, it will get a negative or zero grade.

Finally, the detected actions are denoted as $\Psi$, where each element of $\Psi$ is a tuple  $(s, e, \bm{c}, \text{grade})$ representing one detected action instance. The tuple defines the start, end, class scores, and confidence of the corresponding detected action, respectively.


The APN predicts action progressions at the frame level, and the profile-matching algorithm only involves some simple numerical computations. These two properties enable our APN framework to conduct fine-grained temporal action detection, i.e., every frame in a test video participates in the detection and can be detected as an action boundary. In our experiments, however, we evaluated 1000 frames evenly sampled from each test video. This strategy can reduce the computation significantly with only a negligible reduction in precision (See Table~\ref{tab: sampling}). Moreover, because the proposed profile-matching algorithm detects actions by searching for temporal intervals with complete action evolution, the incomplete action instances in the test videos can be filtered out.

\section{Experiments}\label{sec: experiments}
In this section, we first introduce the datasets and performance metrics, followed by the implementation details. After that, we provide empirical analyses of the proposed APN. Specifically, we compare the performance of using different ordinal regression methods in the APN, compare the APN's performance on different types of actions, and investigate the impact of temporal duration and stride on the APN. Finally, we compare the APN with some state-of-the-art methods.

\subsection{Datasets and Performance Metrics}
\subsubsection{Datasets}
\textbf{THUMOS14} is a benchmark dataset for temporal action detection. It contains videos of 20 classes of human actions with temporal background. The dataset contains 2756 trimmed videos in the training set, 200 untrimmed videos in the validation set, and 212 untrimmed videos in the test set. Almost all untrimmed videos in THUMOS14 contain multiple action instances, which may belong to different classes. Totally, there are 3007 action instances in the 200 validation untrimmed videos and 3358 action instances in the 212 test videos.
\\
\textbf{DFMAD70} is a part of the DCD dataset \cite{Rim2019} for children behavior research. DFMAD70 comprises 63 untrimmed long videos of first-person view, of which 50 are for training and 13 are for testing. Each video contains eight (4+2+2) complete and five (2+2+1) incomplete action instances of three action classes. This dataset has two characteristics: (1) Strict annotations: only complete action instances have frame-level annotations. In contrast, many annotations in THUMOS14 are of incomplete actions or contain multiple action instances; (2) Long duration: the average duration of videos and actions are 676 seconds and 31 seconds, respectively, both of which are much longer than those in other datasets.

\subsubsection{Metrics}\label{metrics}
The mean absolute error \textbf{(MAE)} was used to measure the APN's performance in predicting action progressions. We always normalize the MAE to the range [0, 100] for better interpretability. For example, if $K=25$ in Eq.~\ref{eq: label equation}, we scaled the MAE by a factor of $4$. As a baseline, in the range of $[0, 100]$, the MAE of random guessing is about 33.33 but the minimum MAE without knowing anything is 25 \cref{first}.

The top-1 accuracy \textbf{(Acc.)} was used to measure the classification performance of the APN, i.e., recognizing the action categories.

The Average Recall under the Average Number of proposals \textbf{(AR@AN)} was used to evaluate the quality of the proposals, i.e., for the task of temporal action proposal generation. It reports the recall rates versus the number of proposals that have been evaluated.

The mean Average Precision \textbf{(mAP)} was used for measuring the performance of temporal action detection. It considers the performance of both classification and localization. The mAP under different thresholds of the intersection of union (IoU), which are set to 0.5 by default, are denoted as mAP@IoU (or @IoU).


\subsection{Implementation Details}

\textit{Pre-training}: We used the weights from the pre-trained models trained on the Kinetics-400  \cite{Carreira_2017_CVPR} and ImageNet \cite{deng2009imagenet} to initialize the weights in our I3D and ResNet50 backbones, respectively. 


\textit{Optimization}: We used the Adam \cite{kingma2018method} optimizer to update the network parameters. The learning rate was fixed to 1e-5 when the ordinal regression uses nominal classification, cost-sensitive, or regression heads; it was set to 1e-4 for the others regression methods. The number of training epochs was fixed to 10.

\textit{Profile-matching}: The parameters in Algorithm~\ref{Alg: action search} were set depending on the situation. The thresholds $T_{start}^{max}$/$T_{end}^{min}$ for screening the start/end frames of actions  were set according to the MAE of predicting the action progressions. We empirically set them to 40 and 60 for THUMOS14 and 20 and 80 for DFMAD70. The minimum number of frames of action $T_{len}^{min}$ was set based on the average action duration in the datasets. In practice, we set $T_{len}^{min}$ to 60 and 600 frames for THUMOS14 and DFMAD70, respectively. Finally, the IoU threshold $T_{IoU}$ used in the NMS was fixed to 0.4 unless otherwise stated.

\subsection{Study on Action Progression Networks}
\textbf{Comparison of Ordinal Regression Methods.} The ordinal regression methods used in the APN affect its ability to predict action progressions. We investigated five different types of ordinal regression methods, whose theoretical details can be found in Section~\ref{sec: ordinal regression methods}.

\begin{figure}[tb!]
	\centering
    \includegraphics[width=0.48\textwidth]{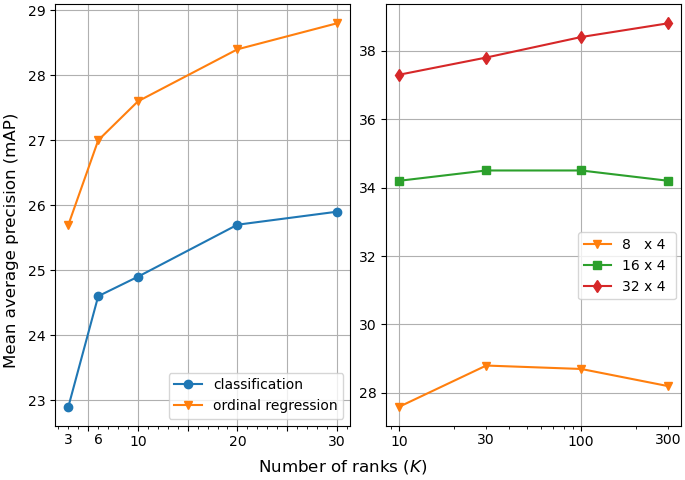}
    \caption{The mean Average Precision (mAP) on THUMOS14 against the number of ranks  of the action progressions ($K$) using the threshold model as the regression method. \textit{\textbf{Left:}} Increasing the number of ranks results in higher mAPs, and encoding with the ordinal regression outperforms the nominal classification. \textit{\textbf{Right:}} The longer duration of the input ($L \times d$ in the legend) the better the performance. The mAP drops as the number of ranks rise to a certain value, which increases with the duration of the input}
    \label{fig: pef vs. ranks}
\end{figure}

The experimental results are summarized in Table~\ref{tab: heads_compare}. We see that different regression methods present different performance and all methods outperform the \textit{nominal classification}, which ignores the order among the stages of the action evolution. Besides, as shown in the \textit{left} part of Fig.~\ref{fig: pef vs. ranks}, the \textit{threshold model} outperforms the nominal classification for various number of ranks in the action progressions. These results demonstrate the merit of using ordinal regression to predict the action progressions instead of classifying the action evolution into three {\it unordered} stages. Due to the superior performance of the threshold model, it was used as the default method in the subsequent experiments unless otherwise stated
\begin{table}[t!]
\caption{Comparison of Ordinal Regression Methods (Section~\ref{sec: ordinal regression methods}) for APN on the THUMOS14 Dataset}
\label{tab: heads_compare}
\centering
\begin{tblr}{
width = .48\textwidth,
colspec = {l X[c] X[c]},
rowsep =1pt,
hline{1, Z} = 1.2pt
}
Ordinal regression method & MAE & mAP\\
\hline
Nominal classification        & 17.85 & 35.8\% \\
Regression                    & 17.62 & 37.6\% \\%
Cost-sensitive classification & 16.99 & 37.5\% \\
Binary decomposition          & 16.21 & 38.5\% \\
Threshold model               & 16.20 & 38.8\% \\
\end{tblr}
\end{table}

\textbf{Impact of the number of Ranks.} In addition to considering the order of evolution stages, another distinct characteristic of action progressions is that they are obtained by dissecting the action evolution into many stages, in contrast to the three stages used in the conventional approachs. We investigated the impact of the number of ranks, i.e., the value of $K$ in Eq.~\ref{eq: label equation}. The experimental results are summarized in Fig.~\ref{fig: pef vs. ranks}. We found that increasing the number ranks for the action progressions leads to higher mAP on temporal action detection. This result further validates the effectiveness of the proposed action progressions. Because of the diminishing return from increasing the $K$, we set 100 as the default value of $K$ in the subsequent experiments unless otherwise stated.

\begin{table}[!tb]
\caption{The mean Average Precision (mAP) of Different Numbers of Sampled Frames on The Test Video}
\label{tab: sampling}
\centering
\begin{tblr}{
width = 0.48\textwidth, 
colspec = {l *{5}{X[c]}},
hline{1, Z} = 1.2pt,
rowsep=1pt
}
\SetCell[r=2]{l} Num. Frames & \SetCell[c=3]{c} THUMOS14 & & & \SetCell[c=3]{c} DFMAD70 & &\\
\cline{2-4} \cline{5-7}
& @0.3 & @0.5 & @0.7 & @0.5 & @0.7 & @0.9 \\
\hline
100   & 37.2 & 22.7 & 7.6  & 77.3 & 35.8 & 1.6 \\ 
200   & 57.3 & 39.1 & 16.7 & 92.8 & 78.4 & 14.3\\
500   & 69.4 & 53.3 & 27.4 & 96.6 & 90.3 & 36.0\\
1000  & 71.0 & 58.3 & 32.8 & 98.9 & 92.9 & 44.2\\
\end{tblr}
\end{table}

\textbf{Impact of the Number of Frames Evaluated}. To investigate the importance of the fine-grained dissection for the precise detection of actions, we conducted experiments to evenly sample different numbers of frames from the test videos, and the results are summarized in Table~\ref{tab: sampling}. We can see that with more sampled frames in the test video, i.e., greater granularity, the APN detects actions more precisely, and it achieves optimal performance when all frames are used for detection, i.e., fully-grained. This result validates our claim that fine-grained temporal contexts are important for precise temporal action detection. We fixed the default granularity to 1000 to achieve a performance close to the fully-grained setting.


\begin{table}[!tb]
\caption{Impact of the Temporal Context of the APN's Input on Performance. $L$: Number of frames. $d$: Stride. Loc: Localization. Cls: Classification}
\label{tab: temporal_size}
\centering
\begin{tblr}{
width=0.48\textwidth,
colspec = {c | X[c] X[c] X[c] X[c]},
rowsep = 1pt,
hline{1,Z} = {1.2pt},
}
& \mbox{Input volume $I_{\tau}$} & Loc &  Cls & Detection\\ 
Row & $L \times d$ & MAE & Acc. & mAP@0.5 \\
\hline
1 & 16 $\times$ 1 & 17.1 & 60.0\% & 30.2\% \\
2 & 16 $\times$ 2 & 15.8 & 65.8\% & 38.2\% \\ %
3 & 16 $\times$ 4 & 15.3 & 70.3\% & 39.1\% \\ %
4 & 16 $\times$ 8 & 14.5 & 71.4\% & 41.2\% \\
\hline
5 & 32 $\times$ 4 & 14.3 & 72.0\% & 44.9\% \\
6 & 64 $\times$ 2 & 14.2 & 71.7\% & 45.0\% \\
7 & 128$\times$ 1 & 15.2 & 64.2\% & 37.7\% \\
\end{tblr}
\end{table}

\textbf{Impact of Temporal Context of the Input.} We tested the APNs at different temporal durations and resolutions. Specifically, we investigated various settings of the number of frames and strides ($L$ and $d$ in Eq.~\ref{eq: input}) for the APNs' input. The experimental results are summarized in Table~\ref{tab: temporal_size}. We have several observations: (1) Comparing the performance of the same number of frames but different strides, thereby different temporal duration, e.g., Rows 1 to 4, we found that longer temporal duration improves the performance in terms of MAE, Acc. and mAP. This is expected because a longer temporal duration contains more information for action detection. (2) Comparing the performances of the same temporal duration but different strides, i.e., Rows 4 to 7, we found that the input volume $32 \times 4$ holds the best balance between performance and the computational cost. The low performance of $128 \times 1$ reflects that there is a lot of redundant information between adjacent frames in the video, and the effectiveness of the pre-trained weights is reduced when the downstream task uses a different temporal stride. We set the default value of $L$ to 32 and $d$ to 4 (i.e., one out of four from 128 consecutive frames) in the subsequent experiments unless stated otherwise.

\begin{figure*}[!t]
	\centering
    \includegraphics[width=1\textwidth]{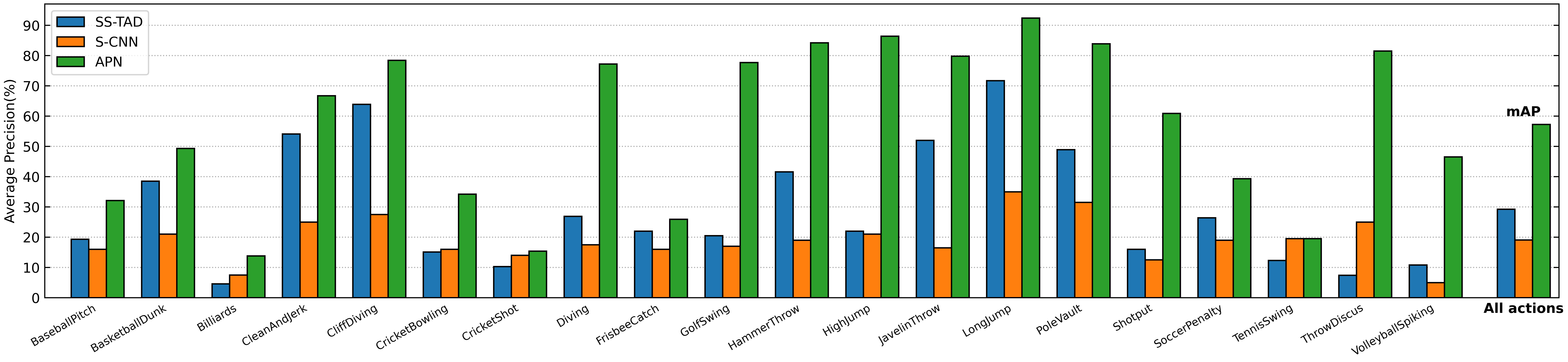}
    \caption{Average precision (\%, @IoU=0.5) for different action classes in THUMOS14. The APN achieves high precision on actions that have a clear evolution pattern (e.g., \textit{Long Jump}) and performs badly on indeterminate actions (e.g., \textit{Billiards}) and actions of short durations (e.g. \textit{Tennis Swing}).}
    \label{fig: every_action}
\end{figure*}

\textbf{Performance on Different Actions}. As introduced in Section~\ref{sec: action progression}, the APN is especially suitable for detecting \textit{progressive} actions. To further investigate this, we present the APN's performance on different action classes in Fig.~\ref{fig: every_action}. There are several observations: (1) The APN outperforms SS-TAD  \cite{sstad_buch_bmvc17} and SCNN \cite{shou2016temporal}, and all three methods perform very differently on detecting actions of different classes. (2) The APN achieves high average precision when detecting actions that have clear evolution patterns, e.g., the APN achieves up to 85\% precision on detecting action "High Jump", whereas the other two methods only achieve about 20\%. Similar situations also occurred in action "Diving", "Golf Swing", and "Throw Discus". In our observations, all these actions have distinct evolution patterns. (3) On the contrary, the APN achieves a poor performance on indeterminate actions, e.g., "Billiards" (13.8\%) and ephemeral actions, e.g., ``Cricket Shot'' (15.4\%) and ``Tennis Swing'' (19.5\%). Despite the difficulty in detecting indeterminate or short-term actions, the APN still outperforms the other two methods. However, the performance gap is narrowed considerably.

\begin{table}[!t]
\caption{Comparison with State-of-the-Art Approaches on THUMOS14 in terms of Temporal Action Proposal Generation (AR@AN)}
\label{Tab: proposal compare}
\centering
\begin{tblr}{
width=0.48\textwidth, 
colspec={l *{5}{X[c]}},
hline{1,Z}=1.2pt
}
Method        & @50 & @100 & @200 & @500 & @1000 \\
\hline
TAG \cite{zhao2017temporal}     & 18.6 & 29.0 & 39.6 & -    & - \\
TURN \cite{gao2017turn}         & 36.5 & 27.8 & 17.8 & -    & -\\
CTAP \cite{gao2018ctap}         & 32.5 & 42.6 & 52.0 & -    & - \\
MGG \cite{liu2019multi}         & 39.9 & 47.8 & 54.7 & 61.4 & 64.1 \\
BSN \cite{Lin_2018_ECCV}        & 37.5 & 46.1 & 53.2 & 60.6 & 64.5\\
BMN \cite{lin2019bmn}           & 39.4 & 47.7 & 54.7 & 62.1 & 65.5 \\
DBG \cite{lin2020fast}          & 37.3 & 46.7 & 54.5 & 62.2 & 66.4 \\
BC-GNN \cite{bai2020boundary}   & 40.5 & 49.6 & 56.3 & 62.8 & 66.6 \\
\hline
APN ($T_{IoU}$=0.8)    & \textbf{45.0} & \textbf{52.9} & \textbf{59.0} & 64.4 & 66.4 \\
APN ($T_{IoU}$=0.85)   & 41.8 & 50.6 & 58.4 & \textbf{64.7} & \textbf{67.8} \\
\end{tblr}
\end{table}

\begin{table*}[!tb]
\caption{Comparison with State-of-the-Art Methods on THUMOS14. E2E: End-to-End; $\dag$: a TSN variant implemented by Xiong \textit{et al.} \cite{xiong2016cuhk}. $\ddag$: models trained on action frames (trimmed videos).}
\label{T14_compare}
\centering
\begin{tblr}{
colspec = {l l l l l l *{5}{c}},
column{6,7,8,9,10,11} = {c},
hline{1,Z} = {1.5pt},
hline{2} = {7-11}{solid, leftpos = -1, rightpos = -1, endpos},
hline{2} = {4-6}{solid, leftpos = -1, rightpos = -1, endpos},
}
Methods &  Modality &  E2E & \SetCell[c=3]{c} Models components & & & \SetCell[c=5]{c} mAP@ & & & & \\
\SetCell[c=3]{l} \textit{Methods based on offline features}  & & & Encoder & Localizer & Classifier & 0.3 & 0.4 & 0.5  & 0.6 & 0.7\\
\hline

TURN \cite{gao2017turn}  & Flow & \ding{55} & TSN$^\dag$ & TURN & SCNN-\textit{cls} \cite{shou2016temporal} & 44.1 & 34.9 & 25.6 & -- & --\\

CTAP \cite{gao2018ctap} & RGB-Flow & \ding{55} & TSN{$^\dag$} & CTAP & SCNN-\textit{cls} & - & - & 29.9 & - & - \\
BSN \cite{Lin_2018_ECCV} & RGB-Flow & \ding{55} & TSN{$^\dag$} & BSN   & SCNN-\textit{cls}                    & 43.1 & 36.3 & 29.4 & 22.4 & 15.0 \\ 
BSN \cite{Lin_2018_ECCV} & RGB-Flow & \ding{55} & TSN{$^\dag$} & BSN   & UNet  \cite{Wang_2017_CVPR}       & 53.5 & 45.0 & 36.9 & 28.4 & 20.0 \\ 
BMN \cite{lin2019bmn}  & RGB-Flow & \ding{55} & TSN{$^\dag$} & BMN & UNet &  56.0 & 47.4 & 38.8 & 29.7 & 20.5 \\
MGG \cite{liu2019multi} & RGB-Flow & \ding{55} & TSN{$^\dag$} & MGG  & UNet                                  & 53.9 & 46.8 & 37.4 & 29.5 & 21.3 \\
DBG \cite{lin2020fast} & RGB-Flow  & \ding{55} & TSN{$^\dag$} & DBG   & UNet                                   & 57.8 & 49.4 & 39.8 & 30.2 & 21.7 \\
G-TAD \cite{xu2020g} & RGB-Flow & \ding{55} & TSN \cite{wang2016temporal} & G-TAD & UNet + \cite{xiong2016cuhk} & 54.5 & 47.6 & 40.2 & 30.8 & 23.4 \\
MR \cite{zhao2020bottom} & RGB-Flow  & \ding{55} & I3D \cite{Carreira_2017_CVPR} & MR & UNet                                   & 53.9 & 50.7 & 45.4 & 38.0 & 28.5\\
P-GCN \cite{zeng2019graph} & RGB-Flow & \ding{55} & I3D & BSN + P-GCN & P-GCN & 63.6 & 57.8 & 49.1 & -    & -    \\
ActionBytes$^\ddag$ \cite{jain2020actionbytes} & RGB-Flow & \ding{55} & I3D & ActionBytes & ActionBytes & 26.1 & 20.3 & 15.5 & - & 3.7 \\
SS-TAD \cite{sstad_buch_bmvc17} & RGB & \ding{55} & C3D \cite{tran2015learning} & SS-TAD & SS-TAD & 45.7 & - & 29.2 & -  & 9.5\\
TadTR \cite{liu2022end}  & RGB-Flow & \ding{55} & I3D & TadTR  & TadTR &  \textbf{74.8} & \textbf{69.1} & \textbf{60.1} & 46.6 & 32.8 \\  

\hline
 \SetCell[c=3]{l} \textit{Methods based on offline proposals}&  &  & Proposer & Detector & &  & &  & & \\
\hline
SCNN \cite{shou2016temporal} & RGB & \ding{55} & SCNN-\textit{prop} \cite{shou2016temporal}  & SCNN & & 36.3 & 28.7 & 19.0 & - & - \\
CDC \cite{Shou_2017_CVPR} & RGB & \ding{55} & SCNN-\textit{prop} & CDC &  & 41.3 & 30.7 & 24.7 & 14.3 & 0.8\\
SSN \cite{zhao2017temporal} & RGB-Flow & \ding{55} & TAG \cite{zhao2017temporal}  & SSN & & 50.6 & 40.8 & 29.1 & -    & - \\

\hline
\SetCell[c=3]{l} \textit{End-to-End methods} & &  & Detector & & &  & &  &  & \\
\hline
SMS \cite{yuan2017temporal} & RGB-Flow & \ding{51} & SMS &  &  & 36.5 & 27.8 & 17.8 & -    & -   \\

APN$^\ddag$ (ours) & RGB-Flow & \ding{51} & APN &  &  & 64.1 & 58.9 & 50.2 & 40.0 & 26.4\\
APN-L$^\ddag$ & RGB-Flow & \ding{51} & APN &  & & 72.4 & 66.7 & 58.3 & \textbf{47.3} & \textbf{33.8} \\
\end{tblr}
\end{table*}

\subsection{Comparison with State-of-the-Art}
Based on the above studies, we found a proper setting for the APN. Then we compared our framework against the state-of-the-art methods in terms of temporal action proposal generation and temporal action detection. 

\subsubsection{Temporal Action Proposal Generation}
We first compared our APN with other methods on the capability of generating temporal action proposals. To generate proposals, we simply ignored the predicted class scores. The comparison was conducted on THUMOS14, and the results are shown in Table~\ref{Tab: proposal compare}. The results show that action progressions are suitable for locating actions. Besides, the APN can adjust the balance between the quality and recall rate of the generated proposals by changing the IoU thresholds in NMS, i.e., $T_{IoU}$ in Algorithm~\ref{Alg: action search}.

\subsubsection{Temporal Action Detection}
We first compared the APN with other methods for temporal action detection on THUMOS14, which are summarized in Table~\ref{T14_compare}. The APN-L stands for an APN variant with a larger backbone: Uniformer \cite{li2022uniformerv2}. We see that our APN achieved competitive performance under the condition that it was trained end-to-end using action frames only. In contrast, the BSN \cite{Lin_2018_ECCV}, as a representative of the bottom-up TAD methods, requires training three different models separately. Similar to the APN, the SMS \cite{yuan2017temporal} utilizes an end-to-end model and the ActionBytes \cite{jain2020actionbytes} was trained on trimmed videos, but our APN achieves much higher precision (58.3\% vs. 17.8\%/15.5\%.).

As mentioned in Section~\ref{sec: action progression}, the APN is designed for detecting progressive actions, especially for long-duration actions and it can avoid detecting incomplete actions. Therefore, we further compared the APN with other methods on the DFMAD70 dataset, in which all annotated actions have clear evolution patterns. Besides, the average duration of actions and videos in DFMAD70 is much longer than those in THUMOS14, and DFMAD70 contains lots of incomplete un-annotated actions.

\begin{figure}[!t]
	\centering
    \includegraphics[width=.48\textwidth]{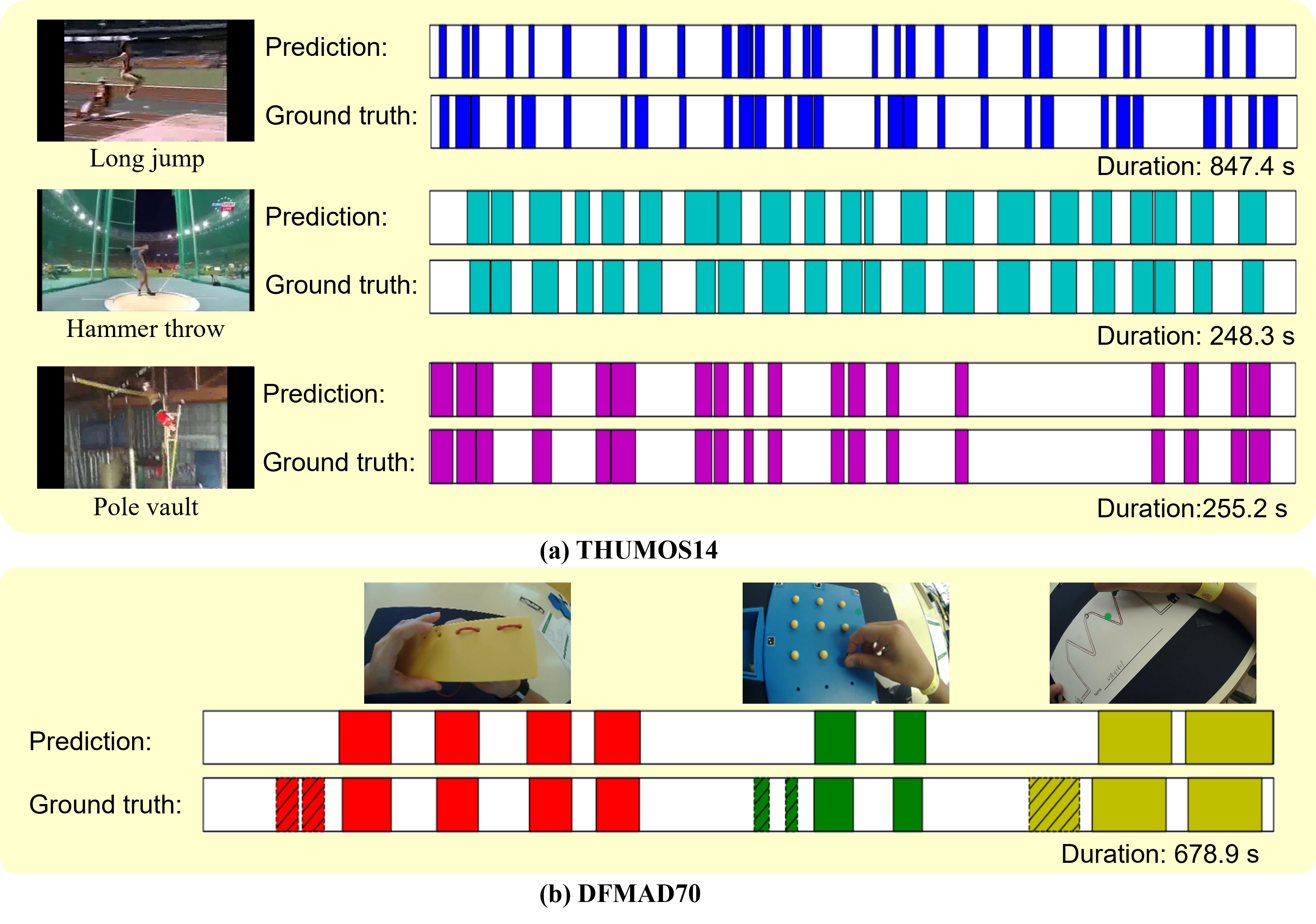}
    \caption{Qualitative results of the APN on (a) THUMOS14 and (b) DFMAD70. Each rectangular block represents one action instance, different colors represent different action classes, and the blocks with hatch stand for incomplete action instances.}
    \label{fig: qualitative}
\end{figure}

The experimental results on DFMAD70 are summarized in Table~\ref{D70_compare}. We compared our method with three approaches: BMN \cite{lin2019bmn}, G-TAD \cite{xu2020g}, and P-GCN \cite{zeng2019graph}. There are several observations: (1) The APN outperforms the three other methods, and the performance gap is larger than in the THUMOS14. For example, the APN achieves up to 98.9\% mAP@IoU=0.5, whereas the others can only achieve about 78\%. (2) The APN achieves 44.2\% mAP when the IoU threshold is up to 0.9, and this can be further improved by using strict thresholds (APN-strict with $T_{start}^{max}=20, T_{end}^{min}=80$). (3) The APN can keep a high detection precision (APN-2D, mAP@0.5=96.0\%) even using a lightweight 2-D backbone (ResNet-50 input with RGB images). The lightweight backbone significantly reduces the running time of the APN. With a 2080Ti NVIDIA GPU, APN-2D solved the TAD in 6277 FPS, more than ten times faster than G-TAD \cite{xu2020g} and SSN \cite{zhao2017temporal}, even though their optical flow computation time was not taken into account. We believe these results show that the APN can detect progressive actions effectively and efficiently. Moreover, the APN avoided the detection of most (98.5\%) of the incomplete action instances in DFMAD70, and an example is shown in Fig.~\ref{fig: qualitative}(b).


%

\begin{table}[!t]
\caption{Comparison of Our Method with Other State-of-the-Art Approaches on the DFMAD70 Dataset}
\label{D70_compare}
\centering
\begin{tblr}{
colspec=l l c c c l,
hline{1,Z} = {1.2pt},
rowsep=1pt}
 & & \SetCell[c=3]{c}{mAP@} & & & \\
\cline{3-5}
Method                 & Modality  & 0.5  & 0.7  & 0.9  & FPS\\
\hline
BMN+P-GCN              & RGB-Flow & 69.1 & 67.8 & 17.7 & $<$ 176\\
G-TAD+SSN              & RGB-Flow & 69.2 & 59.5 & 17.4 & $<$ 443\\
G-TAD+P-GCN            & RGB-Flow & 77.9 & 58.1 & 11.8 & $<$ 175\\
\hline
APN  & RGB-Flow        & \textbf{98.9} & \textbf{92.9} & 44.2 & $<$ 456\\
APN-strict             & RGB-Flow & 87.3 & 82.4 & \textbf{48.5} & $<$ 460\\
APN-2D                 & RGB      & 96.0 & 85.7 & 24.1 & \textbf{6277}\\
\end{tblr}
\end{table}

\subsection{Qualitative Results}

To have a clear observation of the performance of APN on temporal action detection, we visualize the detection results and ground truths of some example videos in Fig.~\ref{fig: qualitative}. Although the test videos are complicated, our APN can detect the temporal boundaries and categories of actions precisely. Moreover, as shown in Fig~\ref{fig: qualitative} (b), the APN ignored the incomplete action instances in the test videos.

\section{Conclusions}
In this paper, we propose to locate actions in videos by analyzing the evolution process of actions. Based on this idea, we advocate the notion of action progression and propose a frame-level temporal action detection model -- Action Progression Network. As demonstrated on action detection datasets, this framework outperforms the existing methods by using a single end-to-end neural network, which only needs to be trained with trimmed videos. Besides, the proposed framework is especially suitable for detecting actions that have clear evolution patterns, and it can avoid the detection of incomplete actions. Moreover, the APN is good at detecting actions of long duration, thanks to its fine-grained dissection on videos. Compared with other temporal action detection approaches, the APN framework brings a brand new workflow rather than a sophisticated neural network model. We believe the APN can get further improved by combining it with other successful techniques.

\ifCLASSOPTIONcaptionsoff
  \newpage
\fi

\printbibliography

\end{document}